\documentclass{article}

\usepackage[final]{nips_2018}

\usepackage[utf8]{inputenc} 
\usepackage[T1]{fontenc}    
\usepackage{hyperref}       
\usepackage{url}            
\usepackage{booktabs}       
\usepackage{amsfonts}       
\usepackage{nicefrac}       
\usepackage{microtype}      
\usepackage{multirow}       
\usepackage{graphicx}       
\usepackage{amsmath}
\setcitestyle{numbers}

\title{\vspace{-4mm}Semi-Supervised Deep Learning for Abnormality Classification in Retinal Images}

\author{
  Bruno Lecouat$^1$ \\
  \And
  Ken Chang$^2$ \\
  \And
  Chuan-Sheng Foo$^1$ \\
  \And
  Balagopal Unnikrishnan\\
  \And
  James M. Brown$^2$ \\
  \And
  Houssam Zenati$^1$\\
  \And
  Andrew Beers$^2$\\
  \And
  Vijay Chandrasekhar$^1$\\
  \AND
  Jayashree Kalpathy-Cramer$^{2*}$\\
  \And
  Pavitra Krishnaswamy$^{1*}$\\
  \And
  \texttt{kalpathy@nmr.mgh.harvard.edu} and \texttt{pavitrak@i2r.a-star.edu.sg}\\
  $^1$ Institute for Infocomm Research, A*STAR, Singapore\\
  $^2$ Athinoula A. Martinos Center for Biomedical Imaging,\\
  Massachusetts General Hospital, Boston, MA, USA
}

\begin{document}

\maketitle
\begin{abstract}
Supervised deep learning algorithms have enabled significant performance gains in medical image classification tasks. But these methods rely on large labeled datasets that require resource-intensive expert annotation. Semi-supervised generative adversarial network (GAN) approaches offer a means to learn from limited labeled data alongside larger unlabeled datasets, but have not been applied to discern fine-scale, sparse or localized features that define medical abnormalities. To overcome these limitations, we propose a patch-based semi-supervised learning approach and evaluate performance on classification of diabetic retinopathy from funduscopic images. Our semi-supervised approach achieves high AUC with just 10-20 labeled training images, and outperforms the supervised baselines by upto 15\% when less than 30\% of the training dataset is labeled.  Further, our method implicitly enables interpretation of the SSL predictions. As this approach enables good accuracy, resolution and interpretability with lower annotation burden, it sets the pathway for scalable applications of deep learning in clinical imaging. 
\end{abstract}
\vspace{-2mm}
\section{Introduction}
\vspace{-2mm}
Deep learning is driving significant advances in automated analysis and interpretation of medical images for applications spanning reconstruction, segmentation, diagnosis, prognosis and treatment response assessment in radiology, dermatology, pathology, oncology and ophthalmology \cite{greenspan2016guest,Esteva2017,Gulshan2016}. Typically, medical imaging studies employ supervised convolutional neural networks (CNNs) that require large datasets annotated by experts to obtain high predictive performance. In many applications, this annotation burden is further exacerbated by the need for multiple annotations to reduce labeling noise \cite{Gulshan2016,Krause2017}.

Semi-supervised deep learning (SSL) algorithms that combine small labeled datasets with larger unlabeled datasets offer a means to address these limitations. Recent works have explored SSL approaches based on generative adversarial networks (GANs), and showed applicability to classifying skin and heart disease \cite{XinYi2018, MadaniISBI2018, MadaniNDH2018}. While these studies have demonstrated feasibility of the GAN approach, they have been limited to low resolution images ($32 \times 32$ to $110 \times 110$ pixels) and to applications where clinically relevant features are present in large parts of the image. However, in many cases, clinical image classification relies on fine features that are only visible in high resolution images and/or are sparsely distributed throughout the image. Further, the ability to interpret classifier predictions would be desirable in practical scenarios.

To address these needs, we frame the semi-supervised medical image classification problem as one of learning from very few labeled images with granular annotations, alongside a larger set of unlabeled images. As an example, we consider patch-level annotations for the labeled set, and propose to perform SSL at the patch level. We then aggregate predictions from individual patches of a given image into an image-level classification without requiring additional annotation. As this approach treats images as composites of finely labeled entities, it can overcome the resolution and interpretation limitations encountered in previous works employing SSL using GANs for medical imaging and computer vision applications \cite{GANMI2018, ladder}.

We demonstrate this approach on the task of classifying abnormalities in retinal fundoscopy images obtained from Diabetic Retinopathy (DR) patients. DR patients are diagnosed based on the presence of fine-scaled, sparse and localized microaneurysms, soft exudates, hard exudates, and hemorrhages, which are hard to annotate. Therefore, computer-aided screening for early detection and intervention \cite{Quellec2017} requires high-resolution images. We show that our GAN-based semi-supervised method can provide accurate classification with far fewer labeled samples than CNN-based supervised methods. Further, we demonstrate that our method can effectively detect fine-scale anomalies and classify spatially sparse abnormalities in an interpretable manner. Finally, we discuss directions to enable translation of semi-supervised deep learning methods for practical use in clinical imaging.
\vspace{-3mm}
\section{Patch-based Semi-Supervised Classification Approach}
\begin{figure}[h]
\noindent 
\begin{centering}
\includegraphics[width=\linewidth]{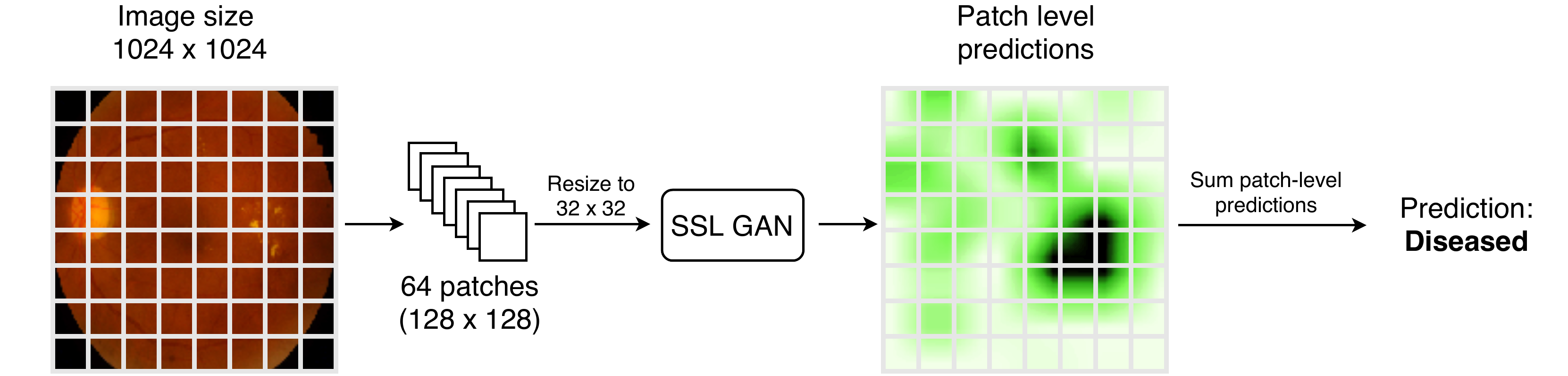}
\par\end{centering}
\caption{Overview of patch-based semi-supervised classification approach.}
\label{fig:overview}
\end{figure}

\vspace{-3mm}
We propose a patch-based semi-supervised classification framework where high-resolution medical images are divided into equal sized patches before being used for training or prediction with a semi-supervised GAN (Figure~\ref{fig:overview}). Predictions on individual patches are then aggregated to produce an image-level prediction. Our patch-based approach enables visualization and localization of salient predictive features by overlaying patch-level predictions onto the input image.
\vspace{-4mm}


\textbf{Semi-supervised GAN (SSL-GAN)}:
GANs \cite{goodfellow2014generative} are a class of deep generative neural network models that have been successful in modeling distributions over natural images \cite{Radford2015}. A typical GAN consists of a generator network $G$ and discriminator network $D$. In the course of training a GAN on a set of (unlabeled) images, the generator learns how to map a low-dimensional set of random variables $Z$ to generate images, while the discriminator learns how to differentiate between those generated images and real images present in the training set.

We build upon the semi-supervised feature-matching GAN framework \cite{Improved}. This method extends the discriminator $D$ to determine the specific class of the image in addition to determining whether the image is real or generated. Formally, suppose we are given a dataset of image patches $\mathcal{I} = \mathcal{I_L} \cup \mathcal{I_U}$ consisting of a labeled set $\mathcal{I}_L = \left\{ (x_i, y_i) \right\}$ and unlabeled set $\mathcal{I}_U = \left\{ x_i \right\}$ of patches; here $x_i$ and $y_i$ are the patches and labels (having $K$ classes) respectively. Then, during training, we simultaneously optimize $D$ and $G$ using stochastic gradient descent, minimizing loss functions $L_D$ for the discriminator and $L_G$ for the generator. $L_D$ is the sum of a loss term on the labeled image subset ($L_{supervised}$) and the vanilla GAN loss ($L_{unsupervised}$) so that
\begin{align*}
L_D &= L_{unsupervised} + L_{supervised} \\
L_{supervised} &= - \mathbb{E}_{(x,y) \sim \mathcal{I}_L} \left[ \log \: p_D (y|x,y<K+1) \right]  \\
L_{unsupervised} &= - \mathbb{E}_{x \sim \mathcal{I}} \left[ \log [1- p_D (y=K+1|x)] \right]  - \mathbb{E}_{x \sim G} \left[ \log [p_D (y=K+1|x)] \right].
\end{align*}
$L_G$, the feature-matching loss designed to encourage generated patches to have similar features to the real patches.
\[
L_G = \left\| \mathbb{E}_{x \sim \mathcal{I} } \left[h(x)\right] - \mathbb{E}_{z \sim p_{z}(z)} \left[h(g(z))\right] \right\|_1
\]
Here, $h(x)$ denotes activations on an intermediate layer of the discriminator. In our experiments, the activation layer after the Networks in Networks (NiN) layers \cite{nin2013} was chosen as intermediate $h(x)$. As the generator and discriminator networks in \cite{Improved} were developed for benchmark natural image datasets, we adapted the networks to the larger image sizes and distinct image statistics of the retinal funduscopy datasets (Experimental Setup, Supplement 1).

\textbf{Image level predictions}: At evaluation time, we applied the semi-supervised GAN to derive patch-level predictions and then pooled these patch-level predictions to form an image-level abnormality score: $\mathrm{score}_i = \sum_{j=1}^{64} \sigma(l_{ij})$ where $l_{ij}$ is the classifier logit of patch $j$ in image $i$. If the image-level abnormality score exceeds a threshold, the classifier predicts that the image is diseased.
\vspace{-3mm}
\section{Experimental setup}
\vspace{-3mm}
\textbf{Dataset and Pre-processing:} We evaluated our SSL-GAN approach on color retinal fundus images from the IDRiD challenge dataset collected at an eye clinic located in Nanded, Maharashtra, India \cite{porwal2018indian}. This dataset contains images from 249 patients (168 healthy, 81 DR). DR patients are diagnosed based on the presence of microaneurysms, soft exudates, hard exudates, and hemorrhages in retinal fundus photographs \cite{Gulshan2016}; the IDRiD dataset contains segmentation masks for each of these abnormalities. 

We randomly split the dataset into Training (n=149), Validation (n=50), and Testing (n=50) cohorts. We resized the original images to $1024 \times 1024$ and normalized them by the maximum intensity value in each image. We then subdivided the normalized image into non-overlapping $128 \times 128$ patches based on a uniform 8x8 grid. To determine patch-level labels, we combined segmentation masks for each of the abnormalities into a single binary mask. Then, we labeled patches with 0 pixels overlapping the mask as healthy; and denoted patches with at least 1 overlapping pixel as diseased. A majority of the patches had very sparse disease features  ($<3\%$ of the patch overlapped abnormality masks). 

\textbf{Baselines:} We compared the semi-supervised approach (SSL-GAN) in the patch-based framework against two supervised baselines: a shallow CNN with architecture similar to the GAN (ConvNet), and a 34-layer residual network (ResNet34, \cite{He2016b}. For both SSL-GAN and supervised baselines, we employed the same method to aggregate patch-level predictions  to produce an image-level classification.

\textbf{Model Training and Evaluation:} The SSL-GAN uses limited labeled data and lots of unlabeled data during training. We sampled labeled images randomly from the training cohort; and used remaining images in the training cohort as unlabeled data. We evaluated the algorithm with different ratios of labeled-to-unlabeled data in the training set. We report mean classification AUCs both at the patch and image levels over 5 random samplings with the associated standard deviations. For appropriate comparisons, we performed supervised baselines with only the limited labeled data samples used in the SSL experiments, and repeated training with 5 different random seeds. We detail model architectures and training hyperparameters in Supplement 1.



\vspace{-3mm}
\section{Results}
\vspace{-4mm}
\textbf{Classification Performance as a Function of Annotation}: We present the semi-supervised classification results, and benchmark against supervised deep learning baselines -- both at the patch and image levels. In particular, we vary the proportion of labeled data in the training set, and evaluate how annotation relates to classification performance across the different methods. 

\begin{table}[htbp!]
  \caption{AUC of Semi-supervised vs. Supervised Learning: Patch-level Classification}
  \label{dr-table1}
  \centering
  \begin{tabular}{llllll}
    \toprule
    Labeled/Total Images:               & 10/149            & 20/149            & 40/149            & 80/149          & 149/149  \\
    \midrule
    SSL-GAN (Patch)             & $\bf{75.8 \pm 2.7}$  & $\bf{79.4 \pm 4.9}$  & $\bf{81.7 \pm 2.9}$ & $\bf{83.3 \pm 1.1}$ & $84.5 \pm 4.7$ \\
    ConvNet (Patch)             & $68.3 \pm 3.9$  & $72.8 \pm 2.4$  & $77.3 \pm 0.14$  & $79.5 \pm 1.8$ & $79.6 \pm 0.22$ \\
    ResNet34 (Patch)        & $52.2 \pm 23.0$  & $61.5 \pm 17.0$  & $73.9 \pm 10.8$  & $81.2 \pm 8.0$ & $\bf{85.1 \pm 2.6}$ \\
    \bottomrule
  \end{tabular}
\end{table}

\begin{table}[htp!]
  \caption{AUC of Semi-supervised vs. Supervised Learning: Image-Level Classification}
  \label{dr-table2}
  \centering
  \begin{tabular}{llllll}
    \toprule
    Labeled/Total Images:               & 10/149            & 20/149            & 40/149            & 80/149          & 149/149  \\
    \midrule
    SSL-GAN (Image)             & $\bf{84.5 \pm 11.5}$ & $\bf{89.0 \pm 10.6}$ & $\bf{94.7 \pm 3.3}$ & $\bf{98.5 \pm 0.6}$ & $\bf{99.1 \pm 0.34}$ \\
    ConvNet (Image)             & $71.1 \pm 12.0$ & $81.4 \pm 0.57$  & $87.4 \pm 2.9$ & $94.5 \pm 0.16$  & $97.8 \pm 1.0$ \\
    ResNet34 (Image)        & $71.3 \pm 16.7$  & $79.4 \pm 15.3$  & $80.5 \pm 5.8$  & $84.2 \pm 2.0$ & $98.9 \pm 1.7$ \\
    \bottomrule
  \end{tabular}
\end{table}

At the patch-level, the SSL-GAN significantly outperforms the supervised baselines. We observe that the SSL-GAN image-level predictions tend to have about 10\% higher AUC than the associated patch-level predictions (Table 1 vs. 2). For the final image-level predictions, the semi-supervised classifier shows significant improvements over supervised baselines, even when less than 10\% of the training dataset is labeled. In particular, when less than 30\% of the training dataset is labeled, the SSL-GAN outperforms CNNs by upto 15\%. We also performed comparisons to a 50-layer residual network trained for direct classification on full-size images (Supplement 2).

\begin{figure}[h]
\begin{centering}
\includegraphics[width=0.85\linewidth]{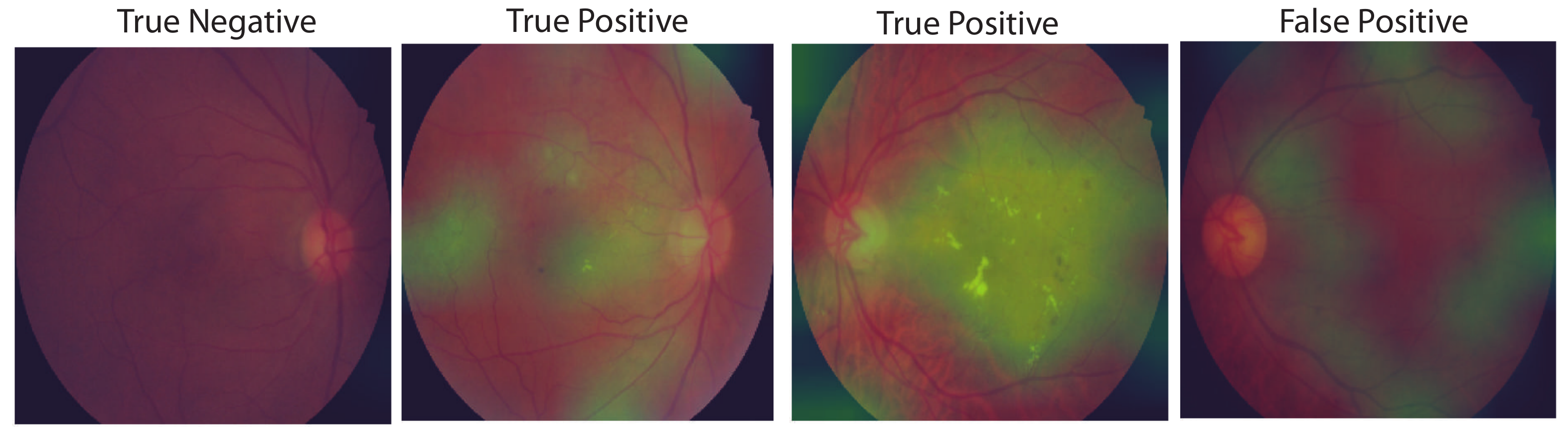}
\par\end{centering}
\caption{Exemplar test-set images with overlay of patch-based abnormality scores predicted by the SSL-GAN. In each case, the image-level classification accuracy is indicated.}
\label{fig:intepretable-gan}
\end{figure}

\textbf{Interpretation of Abnormality Predictions}: To interpret the SSL classification results, we overlaid the localized patch-level abnormality scores spatially onto the image; and smoothed the resulting visualization with a Gaussian blur. Figure~\ref{fig:intepretable-gan} shows some example testing results obtained from an SSL-GAN trained with 20 labeled images. The predictions are clinically meaningful: qualitative comparisons with ground truth segmentation masks suggest that the method can accurately detect exudates and hemorrhages, although there are some misses at the peripheral patches. We quantitatively compared the resulting localization masks against the ground truth annotations, and found that the SSL-GAN had an AUC gain of 16.30\% over the CNN baselines. 

\vspace{-2mm}

\vspace{-2mm}

Taken together, these results suggest that our semi-supervised GAN approach can provide significant improvements over supervised baselines, maintain classification accuracy with large reductions in labeling burden, and enable localization for better interpretation of results. We performed a preliminary test to explore how the models trained on the IDRiD data generalize to an independent test on the Kaggle Diabetic Retinopathy dataset. The SSL-GAN had an AUC of 64\% against 47\% on the supervised CNN baselines, suggesting that the semi-supervised models could exhibit greater capacity to adapt and generalize to varying dataset, class distribution and cohort characteristics.
\vspace{-2mm}
\section{Discussion}
\vspace{-2mm}
We have proposed a patch-based approach to extend SSL with GANs to high-resolution scenarios typical of medical applications. Our approach leverages granular annotations on a minimal dataset, and offers an effective, efficient alternative to supervised approaches that require coarse annotation for large datasets. To the best of our knowledge, this is the first report employing GANs for semi-supervised  classification of fine-scale sparse abnormalities in images. Further, as our semi-supervised classifier produces patch-based predictions, it also implicitly provides a valuable means to interpret the image-level classification results. As such, our work demonstrates that it is possible to use GAN-based semi-supervised deep learning to concomitantly reduce annotation burden, obtain accurate classifications, maintain desirable resolutions and enable interpretation of predictions. This has implications for practical systems focused on scaling applications of deep learning in cross-sectional and multi-dimensional clinical imaging applications.

Although we demonstrated feasibility on retinal image classification, our approach generalizes to a range of classification tasks involving high spatial resolution images and/or sparse anomalous features. Example applications include digital pathology with gigapixel whole-slide images \cite{saltz2018spatial}, cancer screening with cross-sectional CT/MRI \cite{NicolasNatMed2018} and severity grading in multiple sclerosis \cite{SpinalMRI2018}. 

Existing SSL methods have been developed on standard computer vision datasets wherein image structure and composition vary significantly with the target labels. However, as medical images typically have more structural similarity and greater redundancy amongst samples, there is need to design new methods for these unique requirements. 

\bibliography{bibliography.bib}
\bibliographystyle{plain}

\medskip

\section*{Acknowledgments}
This project was supported by funding from the Deep Learning 2.0 program at A*STAR, Singapore, and a training grant from the US National Institute of Biomedical Imaging and Bioengineering (NIBIB, 5T32EB1680). The content is solely the responsibility of the authors and does not necessarily represent the official views of the funders.

\newpage
\appendix
\addcontentsline{toc}{section}{Appendices}
\section*{Appendices}
\section{Architecture and Hyperparameters}

The semi-supervised network architectures are detailed in Tables \ref{discriminator} and \ref{generator}. Briefly, the discriminator comprises a 10-layer convolutional neural network with dropout and weight normalization; and the generator comprises a 5-layer convolutional neural network with batch normalization. For semi-supervised learning and the associated CNN, we downsampled the patches to $32 \times 32$. We augmented the data during training by performing random cropping and flipping of the input training images. We used an exponential moving average of the parameters for inference on the testing set. We used the validation datasets to determine the model hyper-parameters (Supplement Table \ref{param}). The hyper parameters are maintained across all experiments. We will release our code in due course.

\begin{table}[h]
\centering
\caption{Discriminator}
\label{discriminator}
\begin{tabular}{c}
\hline
\multicolumn{1}{c}{conv-large DR}                                      \\\hline
\multicolumn{1}{c}{32$\times$32$\times$3 images}               \\ 
\multicolumn{1}{c}{dropout, $p=0.2$}   \\
\multicolumn{1}{c}{3$\times$3 conv. weightnorm 96 lReLU}              \\
\multicolumn{1}{c}{3$\times$3 conv. weightnorm 96 lReLU}              \\
\multicolumn{1}{c}{3$\times$3 conv. weightnorm 96 lReLU stride=2}     \\ 
\multicolumn{1}{c}{dropout, $p=0.5$}                                                                                                          \\ 
\multicolumn{1}{c}{3$\times$3 conv. weightnorm 192 lReLU}              \\
\multicolumn{1}{c}{3$\times$3 conv. weightnorm 192 lReLU}             \\
\multicolumn{1}{c}{3$\times$3 conv. weightnorm 192 lReLU stride=2}    \\ 
\multicolumn{1}{c}{dropout, $p=0.5$}                                                                                                          \\ 
\multicolumn{1}{l}{3$\times$3 conv. w-tnorm 192 lReLU pad=0 stride=2}  \\
\multicolumn{1}{c}{3$\times$3 conv. w-tnorm 192 lReLU pad=0 stride=2}  \\
\multicolumn{1}{c}{NiN weightnorm 192 lReLU}                         \\
\multicolumn{1}{c}{NiN weightnorm 192 lReLU}                           \\ 
\multicolumn{1}{c}{global-pool}                                                                                                               \\
\multicolumn{1}{c}{dense weightnorm 10}                                                                                                       \\ \hline
\end{tabular}
\end{table}

\begin{table}[h]
\centering
\caption{Generator}
\label{generator}
\begin{tabular}{c}
\hline
\multicolumn{1}{c}{DR}                                            \\ \hline
\multicolumn{1}{c}{latent space 100 (uniform noise)}                                                             \\
\multicolumn{1}{c}{dense 6 $\times$ 6 $\times$ 512 batchnorm ReLU}                                               \\
\multicolumn{1}{c}{5$\times$5 conv.T 256 batchnorm ReLU stride=2}                                                \\ 
\multicolumn{1}{c}{5$\times$5 conv.T 128 batchnorm ReLU stride=2}                                                \\
\multicolumn{1}{c}{5$\times$5 conv.T 128 batchnorm ReLU stride=2}                                                \\ 
\multicolumn{1}{c}{5$\times$5 conv.T 3 weightnorm tanh  stride=2}\\ \hline
\end{tabular}
\end{table}

\newpage
\begin{table}[h]
\centering
\caption{Hyperparameters resnet34}
\label{paramCNN}
\begin{tabular}{ll}
\hline
Hyper-parameter                  & \multicolumn{1}{c}{DR}      \\ \hline
Batch size                       & \multicolumn{1}{c}{32}  \\
Optimizer                        & \multicolumn{1}{c}{ADAM ($\alpha=1*10^{-5}\beta_1=0.9$)}       \\
Epoch                            & \multicolumn{1}{c}{early stopping with patience of 20} \\
Loss                         & binary crossentropy balanced class-weighting \\
Learning rate decay            &  \multicolumn{1}{c}{reduce LR to 10\% of its value with a patience of 10}  \\
Augmentation                     & \begin{tabular}[c]{@{}l@{}}0-180 degree rotation, L-R flipping, U-D flipping\\ 0-30\% horizontal shift, 0-30\% vertical shift\end{tabular}
\end{tabular}
\end{table}

\begin{table}[h]
\centering
\caption{Hyperparameters semi-supervised GAN}
\label{param}
 \begin{tabular}{ll}
\hline
Hyper-parameter                  & \multicolumn{1}{c}{DR}       \\ \hline
Epoch                            & \multicolumn{1}{c}{1200} \\
Batch size                       & \multicolumn{1}{c}{100}                       \\
Leaky ReLU slope                 & \multicolumn{1}{c}{0.2}  \\
Exp. moving average decay             & \multicolumn{1}{c}{0.999}                  \\
Learning rate decay            &  \multicolumn{1}{c}{linear to 0 after 1000 epochs} \\
Optimizer  & \multicolumn{1}{c}{ADAM ($\alpha=3*10^{-4}, \beta_1=0.5$)} \\
Weight initialization            & \multicolumn{1}{c}{Isotropic gaussian ($\mu = 0, \sigma = 0.05$)} \\
Bias initialization              & \multicolumn{1}{c}{Constant (0)}                                 \\ \hline
\end{tabular}
\end{table}

\section{Image Level Classification Results}
We present AUCs obtained from a direct fully supervised image-level classification (Table 5). These results serve as a baseline to assess how the use of finer patch-level annotations and aggregation of patch-level predictions affects supervised image classification.

\begin{table}[h]
  \caption{AUC of Semisupervised vs. Supervised Learning: Direct Image-Level Classification}
  \label{dr-table3}
  \centering
  \begin{tabular}{llllll}
    \toprule
    Labeled/Total Images               & 10/149            & 20/149            & 40/149            & 80/149          & 149/149  \\
    \midrule
    Pretrained (Image)      & $85.8 \pm 20.0$ & $98.2 \pm 0.8$ & $98.8 \pm 1.0$ & $99.2 \pm 0.5$ & $98.8.0 \pm 0.8$ \\
    ResNet50 (Image)        & $56.8\pm 15.8$ & $65 \pm 17.11$ & $67.8 \pm 15.4$ & $76.4 \pm 5.9$ & $82.2 \pm 4.0$ \\
    \bottomrule
  \end{tabular}
\end{table}

We note that pre-training reuses weights from networks that are trained on general scene databases (e.g., ImageNet, CIFAR-10), and produces good results even with low numbers of labeled images. However, it must be noted that such workarounds carry the risk of neglecting critical clinical information, reducing interpretability, and do not apply to common clinical scenarios involving cross-sectional (3D) imaging or video data. 




\section{Future Directions}
Our approach enhances and adapts semi-supervised deep learning to the unique challenges of medical datasets. Future work on the translational front will focus on moving towards practical retinal image classification systems. On the methodological front, our study suggests directions towards improved semi-supervised deep learning methods.

\textbf{Retinal Image Classification}: The current method still shows a number of false negatives at patch level. Employing overlapping patches and/or other unsupervised mechanisms to pool patch-level predictions for image-level classification might help to improve sensitivity. Further, we used consensus labels as ground truth in this study, and will explore the effect of the inter-rater variability on the semi-supervised learning process, especially as the number of labeled samples reduces. Finally, we will investigate the effects of varying patch size, threshold to translate pixel-level labels into patch-level labels, and  benchmark SSL performance for these variations of finer labeling against the naive coarse image-level labeling approach. 


\textbf{Semi-Supervised Deep Learning Methods}: One limitation of the present work is that there is no theoretical guideline for the number of samples to label across the different classes or recommendations for which samples to label. Future work should focus on actively informing the selection of the labeled sample subsets. Further, developing capability to effectively ``pre-train'' GAN-based semi-supervised models and evolving a consensus on comparisons between semi-supervised methods and supervised baselines will be necessary to ground practical translation studies for clinical applications. Comparative evaluations across semi-supervised deep learning approaches and deeper supervised CNNs for a range of medical image classification tasks are needed to address these issues systematically.

\end{document}